\documentclass[fleqn,11pt]{wlscirep}
\usepackage[utf8]{inputenc}
\usepackage[T1]{fontenc}
\usepackage{longtable}
\usepackage{booktabs}
\usepackage{ragged2e} 
\usepackage{array}
\usepackage{caption}
\usepackage{multirow}
\usepackage{marvosym}
\usepackage{soul}
\usepackage{xcolor}

\title{Embodied Science: Closing the Discovery Loop with Agentic Embodied AI}

\author[1,2]{Xiang Zhuang}
\author[1]{Chenyi Zhou}
\author[1]{Kehua Feng}
\author[1]{Zhihui Zhu}
\author[3]{Yunfan Gao}
\author[3]{Yijie Zhong}
\author[1]{Yichi Zhang}
\author[1]{Junjie Huang}
\author[1\Letter]{Keyan Ding}
\author[2\Letter]{Lei Bai}
\author[3\Letter]{Haofen Wang}
\author[1\Letter]{Qiang Zhang}
\author[1\Letter]{Huajun Chen}
\affil[1]{Zhejiang University, Hangzhou, China}
\affil[2]{Shanghai Artificial
Intelligence Laboratory, Shanghai, China}
\affil[3]{Tongji University, Shanghai, China}

\affil[\Letter]
{
baisanshi@gmail.com,
haofen.wang@tongji.edu.cn, \{dingkeyan,qiang.zhang.cs,huajunsir\}@zju.edu.cn}

\begin{abstract}

Artificial intelligence has demonstrated remarkable capability in predicting scientific properties, yet scientific discovery remains an inherently physical, long-horizon pursuit governed by experimental cycles. Most current computational approaches are misaligned with this reality, framing discovery as isolated, task-specific predictions rather than continuous interaction with the physical world. 
{Here, we argue for embodied science, a paradigm that reframes scientific discovery as a closed loop tightly coupling agentic reasoning with physical execution.}
We propose a unified \textbf{P}erception–\textbf{L}anguage–\textbf{A}ction–\textbf{D}iscovery (\textbf{PLAD}) framework,
wherein embodied agents perceive experimental environments, reason over scientific knowledge, execute physical interventions, and internalize outcomes to drive subsequent exploration.
By grounding computational reasoning in robust physical feedback, this approach bridges the gap between digital prediction and empirical validation, offering a roadmap for autonomous discovery systems in the life and chemical sciences.

\end{abstract}
\begin{document}

\flushbottom
\maketitle
%
%
\thispagestyle{empty}


\section{Introduction}

Artificial intelligence is reshaping how scientific knowledge is produced and acted upon~\cite{wang2023scientific}. Over the past decade, data-driven models have delivered striking advances in problems once considered intractable, from accurate protein structure prediction~\cite{jumper2021highly,baek2021accurate} to learned models for molecular property prediction~\cite{wu2018moleculenet,yu2023enzyme,fang2023knowledge}, generative design~\cite{krishna2024generalized,zeni2025generative}, and synthesis planning~\cite{han2024retrosynthesis}. These successes, together with the rise of foundation models that unify representations across modalities~\cite{wang2026multimodal,bai2025interns1scientificmultimodalfoundation,zhuang2025advancing}, have begun to shift AI for Science (AI4S) from a collection of specialized predictors toward more general-purpose scientific engines~\cite{xu2025probing}.

Yet a central tension remains: \emph{scientific discovery is not a single-shot inference problem}. 
{Breakthroughs typically emerge from long-horizon, iterative interaction with the physical world, through a sustained loop of hypothesis formation, experimental design, execution under real constraints, analysis, and model revision~\cite{kuhn1970structure,popper2005logic}}.
Even when a predictive model is excellent, the discovery process can stall if the system cannot decide what to do next, cannot perceive the signals from instruments, or cannot reliably translate decisions into executable laboratory operations.

Recent efforts make this mismatch visible in the form of a bifurcated landscape. On one side, large language
model (LLM)-based agents~\cite{gao2024empowering, DBLP:journals/corr/abs-2508-14111,naveed2025comprehensive,achiam2023gpt,li2026agentic} expand cognitive scope through language-mediated reasoning, planning, and tool use, often translating high-level scientific intent into experimental plans, code, and workflows. On the other side,
automated and robotic laboratories~\cite{tom2024self, canty2025science} demonstrate reliable embodied execution, enabling sustained experimentation
and closed-loop optimization within well-defined experimental spaces.
This split reflects broader empirical evidence that AI’s currently attributed scientific impact remains concentrated in cognitive augmentation, especially in data processing and pattern recognition, whereas the next critical step is to expand sensory and experimental capacity to enable the search for and acquisition of new forms of evidence beyond ``standing'' datasets~\cite{hao2026artificial}.
Crucially, this split is not an incidental integration gap that will vanish by ``connecting'' modules. Each line optimizes a different projection of discovery under the prevailing framing: cognition is powerful but weakly grounded in instrument-level evidence and physical constraints, whereas execution is robust but often optimized around predefined objectives and procedural boundaries.
Without reframing discovery as an end-to-end closed loop, incremental scaling (such as stronger planners, better robots, or larger models) tends to reproduce fragmentation rather than resolve it.

\begin{figure*}[t]
    \centering
    \includegraphics[width=\textwidth]{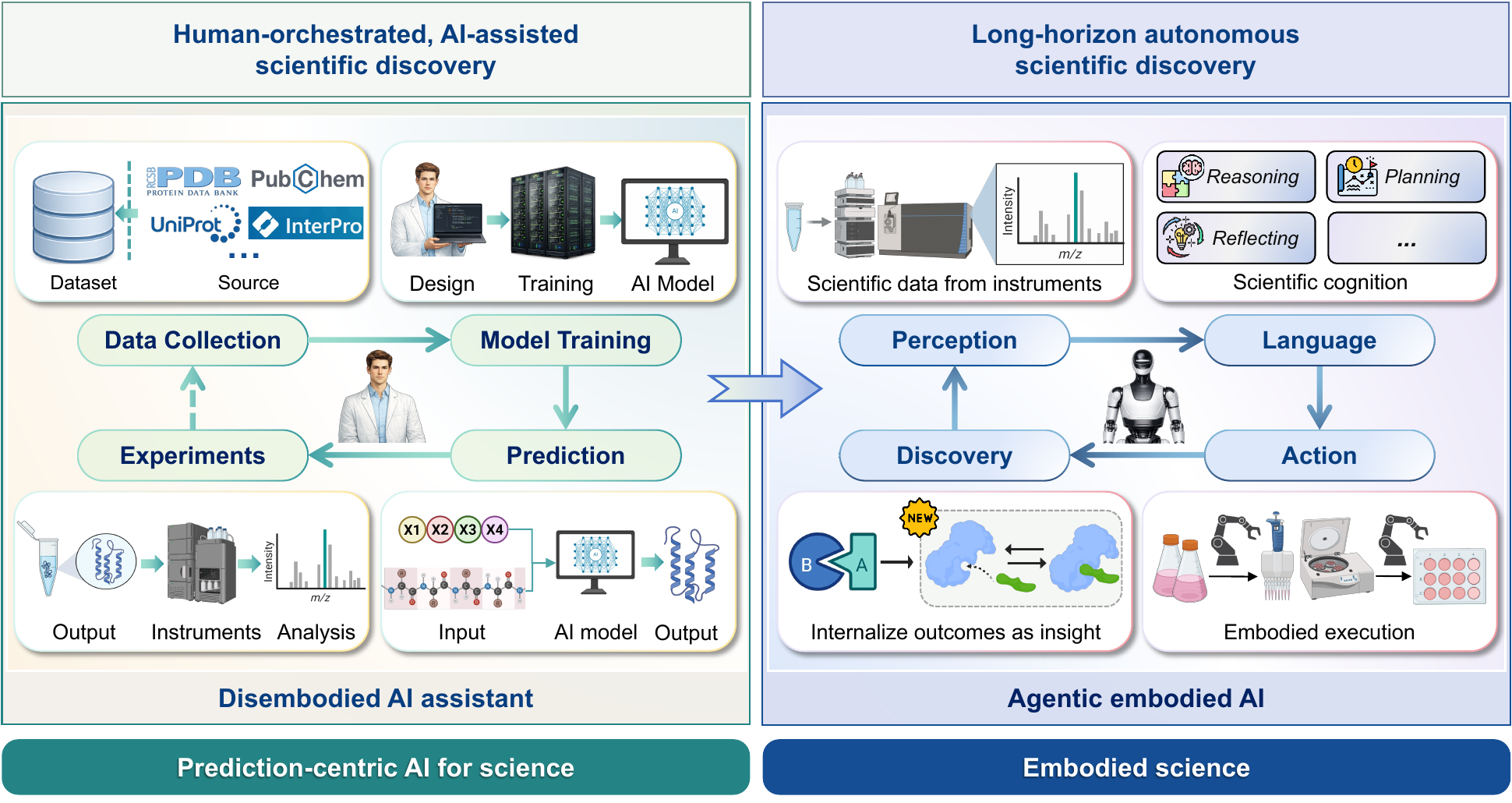}
\caption{\textbf{From prediction-centric AI for science to embodied science.}
Left: the dominant AI4S workflow is human-orchestrated. Experts curate data sources into datasets, train task-specific models, and use predictions to inform the next experimental step; execution remains human-managed, yielding a loosely coupled loop across stages. 
Right: \emph{embodied science} reframes discovery as a closed-loop process interaction with the physical world. \emph{Agentic embodied AI} provides the enabling system layer that operationalizes this paradigm by integrating Perception -- Language -- Action -- Discovery (PLAD): it grounds cognition in instrument-derived evidence (Perception), conducts reasoning and planning (Language), carries out embodied experimentation (Action), and internalizes outcomes into scientific insight (Discovery). The system can sustain \emph{long-horizon autonomous scientific discovery}, executing iterative cycles beyond task-bounded, single-shot assistance.}

    \label{fig:intro}
\end{figure*}

Thus, autonomy in scientific discovery is a property of the coupled system, and realizing the capacity expansion required for continued exploration~\cite{hao2026artificial} collapses when any of the three structural requirements is externalized:
\textbf{perception}, \textbf{cognition}, and \textbf{action}.
\textbf{At the level of perception}, scientific evidence is generated by instruments, not datasets: raw spectra, chromatograms, microscopy streams, sensor logs, calibration traces, and meta-data that capture drift, failure, and context. Models trained on curated tables often lack the ability to parse these multimodal, imperfect signals and to perform goal-directed sensing (e.g., adaptively zoom, re-measure, recalibrate, or switch modalities when anomalies occur).
\textbf{At the level of cognition}, most AI4S systems optimize well-defined tasks (predict a structure, rank candidates, regress a property), but long-horizon discovery requires persistent goal management, experimental reasoning under uncertainty, and planning over contingencies and costs~\cite{DBLP:conf/iclr/YaoZYDSN023}. It demands not only selecting the next experiment, but also deciding \emph{what to measure}, \emph{when to intervene}, and \emph{how to revise hypotheses} as evidence accumulates.
\textbf{At the level of action}, discovery hinges on interventions in the world~\cite{DBLP:journals/corr/abs-2506-22355}. Many AI4S pipelines terminate at candidate suggestions, while real laboratories require precise, verifiable actions: manipulating reagents, configuring instruments, executing protocols, ensuring safety constraints, and recovering from errors. Without robust action grounding, “recommendations” cannot mature into discoveries.

Here, we argue for \textbf{embodied science} as a foundational paradigm for long-horizon autonomous discovery.
In embodied science, AI does not merely analyze data or recommend actions; it expands sensory and experimental capacity by participating directly in experimental workflows as part of a closed loop that couples perception of instrument signals, knowledge-grounded reasoning, and physical intervention.
Scientific progress, under this paradigm, emerges from continuous engagement with real experimental environments rather than from one-off computation over static datasets. This perspective reframes AI-driven discovery as an exploration-driven process, in which hypotheses, experimental strategies, and operational behaviors co-evolve through repeated interaction with the physical world.

\section{Scoping Embodied Science and Agentic Embodied AI}

AI-driven discovery is transitioning from tool-based augmentation to system-level reconfiguration of the scientific method.
To avoid terminology drift, we scope two concepts that organize this Perspective, \textbf{Embodied Science} and \textbf{Agentic Embodied AI}, and we define \textbf{long-horizon autonomous scientific discovery} as the operational outcome they aim to enable.

\subsection{Embodied Science}

\textbf{Definition.}
We use \emph{Embodied Science} to denote a paradigm in which discovery is treated as an embodied, long-horizon, closed-loop process.
AI is integrated into real experimental workflows and operates across the complete cycle of discovery by perceiving instrument-generated signals, reasoning with scientific knowledge, and executing laboratory interventions. Scientific progress therefore arises from sustained interaction with the physical world, rather than from isolated computation over static datasets.

\textbf{Why embodiment matters in scientific exploration.}
Embodiment, in this view, is not laboratory automation in disguise.
It is what makes AI-driven discovery actionable: plans are realized as physical interventions in the laboratory, and their consequences are returned as instrument-grounded feedback.
Rather than mechanically executing fixed protocols, embodied capabilities enable systems to carry intent into the real world, observe what actually happens, and adapt subsequent decisions accordingly.

\subsection{Agentic Embodied AI}

Embodied Science defines \emph{what} kind of discovery process is needed; \textbf{Agentic Embodied AI} specifies \emph{what kind of AI system} can realize it.

\textbf{Definition.}
Agentic Embodied AI is a persistent cyber--physical scientific agent that couples (i) scientific cognition, (ii) experimental perception, and (iii) laboratory action within a single closed-loop controller, operating under explicit feasibility and safety constraints.

Three properties are essential:
\textbf{(1) Agentic autonomy:} the ability to manage goals over extended horizons, plan under uncertainty and cost, and revise strategies in response to outcomes;
\textbf{(2) Embodiment in the experimental loop:} interfaces to raw instrument streams, operational state, and actuation primitives to ground reasoning in laboratory reality;
\textbf{(3) Long-horizon persistence:} memory, provenance, monitoring, and recovery behaviors that maintain continuity across cycles rather than treating each run as a standalone episode.

\subsection{Long-Horizon Autonomous Scientific Discovery}

Because “autonomy” is often used loosely, we adopt an operational criterion:

\textbf{Operational criterion.}
A system demonstrates \emph{long-horizon autonomous discovery} if it can sustain multiple end-to-end discovery cycles---hypothesis $\rightarrow$ experiment design $\rightarrow$ physical execution $\rightarrow$ interpretation $\rightarrow$ revision---over extended time spans with minimal human intervention, while maintaining reproducibility, provenance, and safety.
This criterion sets a higher bar than most current demonstrations: it requires the loop to remain closed beyond a single episode, sustaining autonomous operation across instrument drift, stochastic outcomes, and accumulating uncertainty, while preserving reproducibility, provenance, and safety.

\section{Analyzing the Current Landscape: Reasoning-Centric and Execution-Centric Paradigms}
Motivated by the need to bridge computational reasoning and physical experimentation, scientific agents and embodied AI have recently emerged as promising routes toward more autonomous scientific discovery. In practice, however, existing efforts have largely bifurcated into two partial realizations: reasoning-centric systems that advance hypothesis generation and in silico exploration, and execution-centric platforms that enable autonomous, high-throughput experimentation within well-defined procedural boundaries.
Each excels at a single component of the discovery process while remaining fundamentally decoupled from the others. Below, we analyze the current landscape and argue that these limitations are structural rather than incremental.

\subsection{Disembodied Scientific Cognition: Reasoning without Physical Action}

Artificial intelligence has long supported scientific research through data-driven modeling and prediction, a trajectory often associated with the Fourth Paradigm of data-intensive science~\cite{hey2009fourth}. In recent years, this role has expanded toward cognition-centric scientific agents that aim to elevate AI from a passive analytical tool to an active reasoning entity. As summarized in Table~\ref{tab:review}, these approaches are unified by a common paradigm: they operate primarily on curated datasets, emphasize language-based scientific reasoning, and remain physically disembodied, with experimental execution and validation performed by humans.

Within this paradigm, a first subclass focuses on task decomposition and planning. Systems such as ChemCrow~\cite{m2024augmenting}, Biomni~\cite{huang2025biomni}, SciToolAgent~\cite{ding2025scitoolagent}, and ToolUniverse~\cite{DBLP:journals/corr/abs-2509-23426} leverage large language models to orchestrate domain-specific tools, dynamically assembling workflows for synthesis planning, reaction optimization, or biomedical analysis. Their strength lies in structuring complex scientific tasks and coordinating heterogeneous computational resources. However, their operational boundary is fundamentally cognitive: experimental actions are abstracted as tool calls, and physical execution remains external.
A second subclass emphasizes hypothesis generation and problem-space exploration. Platforms such as Virtual Lab~\cite{swanson2025virtual}, Robin~\cite{DBLP:journals/corr/abs-2505-13400}, and AI Co-scientist~\cite{penades2025ai} aim to emulate aspects of collaborative scientific reasoning by coordinating evidence gathering, hypothesis refinement, and iterative discussion across specified research problems. These systems move beyond task execution toward exploratory inquiry, yet their exploration remains confined to literature, databases, and simulations. Hypotheses are evaluated through manual evaluation by human experts rather than autonomous experimental falsification, leaving the core scientific loop incomplete.
A third line of work frames discovery as iterative search and optimization in silico. Systems such as AlphaEvolve~\cite{DBLP:journals/corr/abs-2506-13131}, DeepScientist~\cite{DBLP:journals/corr/abs-2509-26603} and InternAgent~\cite{team2025internagent, feng2026internagent} formalize scientific progress as a feedback-driven process in which hypotheses or programs are iteratively refined based on computational evaluation. While this paradigm introduces an explicit notion of iteration and feedback, the feedback signal is constrained to the execution of computational algorithms. Lacking physical embodiment and the ability to interact with the material world, these systems struggle to generalize across broader scientific domains, often producing candidates that fail to translate into physical experimental success.
Finally, some efforts toward end-to-end research in silico, exemplified by systems such as AI Scientist~\cite{DBLP:journals/corr/abs-2408-06292} and Kosmos~\cite{DBLP:journals/corr/abs-2511-02824}, extend cognition-centric agents to include automated manuscript writing and reporting. 
Although these systems appear to close the research loop across problem formulation, methodological design, experimental execution, and writing, their scope remains confined to computational domains.
Without physical embodiment, they remain incapable of engaging with tangible experimental environments, leaving the loop fundamentally decoupled from the complexities of the material world.

Despite their diversity, reasoning-centric scientific agents share a fundamental structural limitation. Without embodied execution in real experimental environments, hypotheses cannot be autonomously tested, falsified, or revised through sustained interaction with the physical world. Feedback is indirect, delayed, or entirely computational, leading to a form of cognitive closure in which reasoning saturates without experimental grounding. Consequently, these approaches struggle to support long-horizon autonomous scientific discovery, which requires repeated, embodied cycles of hypothesis testing and revision.

\begin{table}[t]
\centering
\caption{Landscape of current AI4S approaches toward autonomous scientific discovery.
Existing efforts cluster into two partial realizations: reasoning-centric but physically disembodied systems, and execution-centric but cognitively shallow platforms. Notably, neither paradigm achieves a sustained coupling between language-level scientific reasoning and embodied experimental action.}
\renewcommand{\arraystretch}{1.2}
\scalebox{0.95}{{
\begin{tabular}{
p{3cm} 
p{6cm}  
p{3.5cm}  
p{4cm}  
}
\toprule
\textbf{Paradigm} & \textbf{Characteristics} & \textbf{Subtypes} & \textbf{Example approaches} \\
\midrule
\multirow{4}{=}{Language-heavy,\\ action-light\\ (physically disembodied)~\cite{gao2024empowering,DBLP:journals/corr/abs-2508-14111}}
&
\multirow{4}{=}{$\bullet$ Strong language-based reasoning, but no coupling to embodied action; \\
$\bullet$ Operate primarily on curated datasets rather than raw signals from instruments; \\
$\bullet$ Remain physically disembodied, with experimental execution carried out by human scientists.
}

& Task decomposition \& planning
& e.g., ChemCrow~\cite{m2024augmenting}; Biomni~\cite{huang2025biomni}; SciToolAgent~\cite{ding2025scitoolagent}; ToolUniverse~\cite{DBLP:journals/corr/abs-2509-23426} \\
\cline{3-4}
& & Hypothesis \& problem exploration
& e.g., Virtual Lab~\cite{swanson2025virtual}; Robin~\cite{DBLP:journals/corr/abs-2505-13400}; AI Co-scientist~\cite{penades2025ai} \\
\cline{3-4}
& & Iterative search \& optimization in silico
& e.g., AlphaEvolve~\cite{DBLP:journals/corr/abs-2506-13131}; DeepScientist~\cite{DBLP:journals/corr/abs-2509-26603}; InternAgent~\cite{team2025internagent,feng2026internagent} \\
\cline{3-4}
& & End-to-end research in silico
& e.g., AI Scientist~\cite{DBLP:journals/corr/abs-2408-06292}; Kosmos~\cite{DBLP:journals/corr/abs-2511-02824} \\
\midrule
\multirow{4}{=}{Action-heavy,\\ language-light\\ (cognitively shallow)~\cite{tom2024self, canty2025science}}
&
\multirow{4}{=}{$\bullet$ Reliable embodied execution, but limited language-level reasoning and hypothesis formation; \\
$\bullet$ Operate directly on raw instrument signals using heuristic decision-making (e.g., Bayesian optimization); \\
$\bullet$ Enable autonomous execution within narrow task scopes and fixed procedural boundaries.
}

& Single-step, instrument-bound
& e.g., automated liquid handlers; robotic pipetting and weighing systems \\
\cline{3-4}
& & Multi-step, protocol-driven automation
& e.g., Chemputer~\cite{steiner2019organic}; FLUID~\cite{kuwahara2025development} \\
\cline{3-4}
& & Closed-loop execution and optimization
& e.g., A-Lab~\cite{szymanski2023autonomous} RoboChem~\cite{slattery2024automated}; CRESt~\cite{zhang2025multimodal} \\
\cline{3-4}
& & Language-instructed experimental execution
& e.g., Coscientist~\cite{boiko2023autonomous}; ChemAgents~\cite{song2025multiagent} \\

\bottomrule
\end{tabular}
}}
\label{tab:review}
\end{table}
\subsection{Execution-Centric Embodiment: Action without Scientific Understanding}
In parallel with advances in cognition-centric scientific agents, embodied automation has made substantial progress toward the physical execution of experiments~\cite{king2009automation}. As characterized in Table~\ref{tab:review}, execution-centric embodied systems operate on raw instrument signals, rely primarily on heuristic or statistical decision-making, and enable direct interaction with the physical world. However, these systems are typically cognitively shallow: while they excel at executing experiments, they lack the capacity for mechanism-aware reasoning and hypothesis-driven inquiry.

At one end of the execution-centric spectrum are single-step, instrument-bound systems. Automated liquid handlers, robotic pipetting platforms, and weighing systems exemplify this class. They automate isolated experimental operations with high precision and repeatability, yet each action is executed independently of the broader scientific context. Consequently, these systems constitute the basic infrastructure of automated experimentation: they robustly perform prescribed single-step actions, yet lack the capacity to analyze outcomes, reason about experimental context, or adapt behavior beyond explicit instructions.
A more advanced, yet still execution-bound subclass extends automation to multi-step, protocol-driven execution. Platforms such as the Chemputer~\cite{steiner2019organic} and FLUID~\cite{steiner2019organic} encode experimental procedures as machine-executable workflows, enabling the automated realization of complex, multi-stage protocols. Compared to single-step systems, this approach extends automation from isolated actions to coordinated, multi-stage experimental execution. Although execution now spans multiple stages, the system does not analyze intermediate outcomes, reason about experimental context, or deviate from the prescribed workflow when unexpected results occur.
At the most integrated end of execution-centric systems are closed-loop, feedback-driven platforms, including A-Lab~\cite{szymanski2023autonomous}, RoboChem~\cite{slattery2024automated}, and CRESt~\cite{zhang2025multimodal}. By integrating robotic execution with online characterization and statistical optimization methods, including Bayesian optimization and evolutionary search, these platforms demonstrate impressive autonomy over extended experimental campaigns, adapting parameters in response to observed outcomes and achieving local optimization in high-dimensional spaces. However, the closed loop operates primarily at the level of numerical feedback rather than scientific representation. When experiments fail or yield anomalous results, adaptation proceeds through parameter adjustment rather than counterfactual reasoning or hypothesis revision.
With the introduction of large language models, robotic systems have begun to blur the boundary between execution-centric automation and cognitive planning, using LLMs to design experimental workflows and drive embodied laboratory systems for physical execution~\cite{boiko2023autonomous,song2025multiagent}. Despite this progress, LLM integration primarily improves the flexibility of workflow design and control, without conferring sustained, long-horizon scientific agency. Experimental outcomes are used to refine individual tasks, leading to episodic, task-bounded reasoning rather than cumulative, discovery-driven iteration.

Despite increasing integration, execution-centric systems face two persistent limitations at the level of embodiment. First, most platforms rely on fixed or rail-mounted manipulators tightly coupled to predefined laboratory layouts, which provide reliability but limit physical flexibility across heterogeneous instruments and reconfigurable environments. Second, the development, calibration, and maintenance of such systems remain labor-intensive, requiring substantial human effort to debug workflows and adapt execution logic to new experimental settings.
To alleviate these constraints, execution-centric autonomy has been extended through mobile embodiments and digital twin–based simulation. Mobile robotics~\cite{burger2020mobile,dai2024autonomous} address the limitation of rigid embodiment by physically interconnecting spatially distributed instruments, thereby expanding the scope of executable workflows beyond fixed workcells. In parallel, digital twin frameworks such as MATTERIX~\cite{darvish2025matterix} target the engineering bottleneck by enabling virtual validation and sim-to-real transfer~\cite{zhao2020sim}, reducing the cost and risk associated with deploying automated experimental workflows. However, both advances operate strictly at the level of execution. They improve flexibility and reliability in how experiments are carried out, but do not enable systems to interpret experimental outcomes, revise scientific hypotheses, or engage in mechanism-aware reasoning. In this sense, mobile robots and digital twins address execution bottlenecks, not scientific cognition.

Consequently, execution-centric embodied systems often function as powerful execution engines rather than as scientists. Their decision-making is guided by predefined objective functions and fixed parameter spaces, limiting their ability to generalize beyond narrowly scoped tasks. Although they may achieve efficient optimization within constrained domains, they do not accumulate transferable scientific insight. Experimentation is treated as a process of executing and scoring trials, rather than as an active inquiry in which hypotheses are formulated, challenged, and revised. This gap between action and understanding gives rise to an illusion of scientific progress: local performance improves, yet discovery remains decoupled from scientific understanding.

\subsection{Why Incremental Scaling Is Insufficient}
The limitations above do not disappear by scaling one side.
More powerful language reasoning does not automatically yield procedural correctness, instrument awareness, or safe execution.
More capable robotics does not automatically yield hypothesis-driven inquiry, evidence integration, or scientific generalization.
Long-horizon autonomy requires a unified system-level coupling between perception, language-level reasoning, embodied action, and cumulative discovery~\cite{XU2026,shi2026knowledge}.
This motivates a closed-loop framework that treats autonomy as an end-to-end property rather than a component upgrade.

\begin{figure*}[t!]
    \centering
    \includegraphics[width=\textwidth]{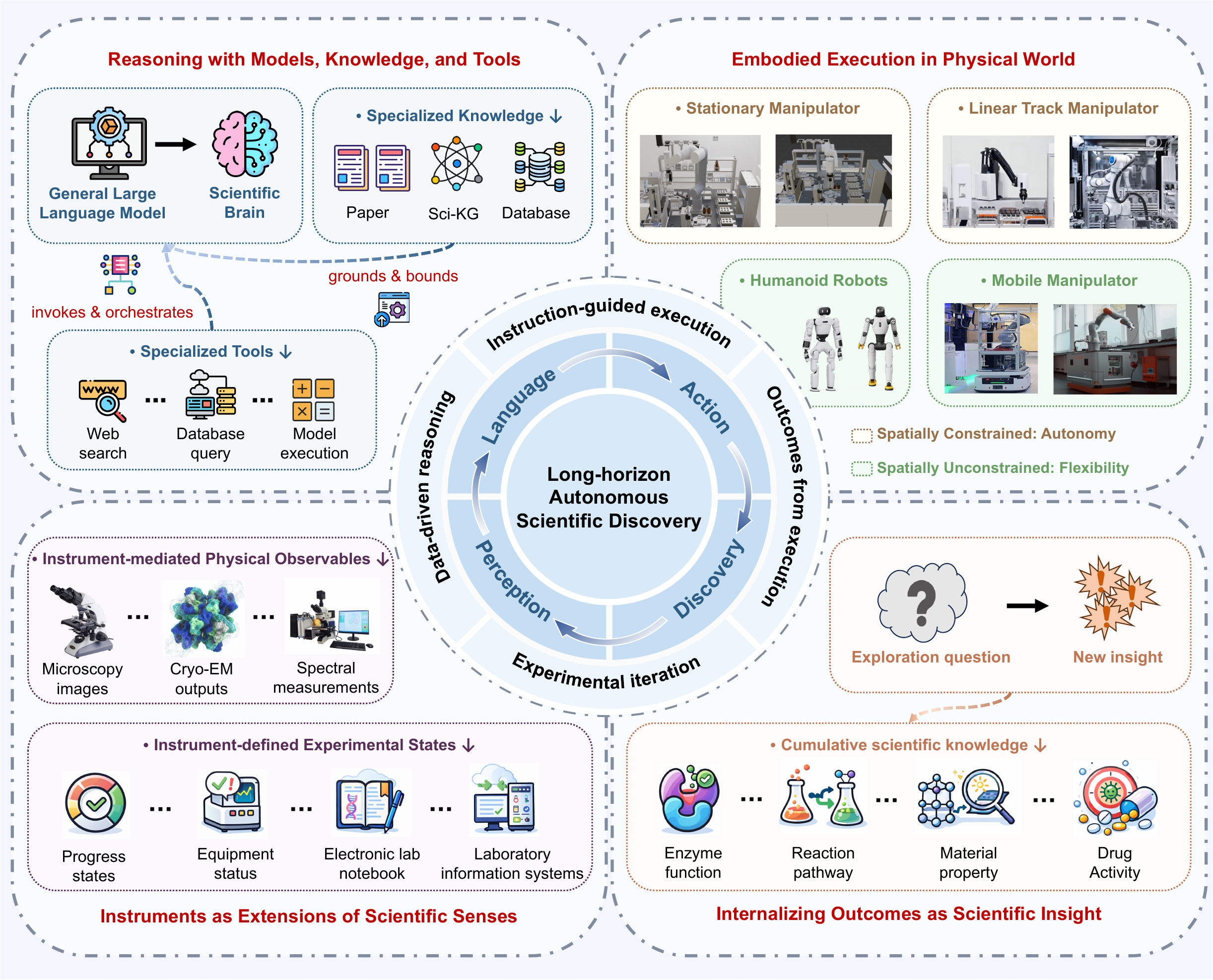}
    \caption{\textbf{The PLAD loop for long-horizon autonolous scientific discovery.}
    \textbf{Perception} transforms raw instrument signals into structured evidence.
    \textbf{Language} integrates foundation models with specialized knowledge and tools to support hypothesis formation, reasoning, and experimental planning.
    \textbf{Action} compiles plans into verifiable laboratory operations.
    \textbf{Discovery} internalizes outcomes into transferable scientific knowledge that shapes subsequent exploration across cycles.}
    \label{fig:framework}
\end{figure*}

\section{PLAD: Towards Closed-Loop Agentic Embodied AI for Scientific Discovery}
We argue that agentic embodied AI represents a critical technological pathway toward long-horizon autonomous scientific discovery. To this end, we propose the Perception–Language–Action–Discovery (PLAD) closed-loop paradigm (Figure~\ref{fig:framework}) as an overarching framework for implementation. Unlike the Vision–Language–Action (VLA)~\cite{zitkovich2023rt} paradigm that underpins general-purpose embodied intelligence, where the primary objective is to understand open environments, generate linguistic descriptions, and execute physical actions. PLAD is explicitly centered on scientific discovery as its core goal. Accordingly, its design is aligned with the distinctive requirements of scientific research across cognitive structures, perceptual targets, and forms of action.

Within PLAD, scientific discovery is modeled as a continuously operating closed-loop process. An agent perceives the experimental environment (Perception), reasons and plans under the support of scientific language and knowledge (Language), executes experiments through embodied actions in real laboratory settings (Action), and internalizes experimental outcomes as new scientific insights (Discovery), which in turn drive subsequent rounds of exploration. Below, we introduce each component of the PLAD paradigm in detail.
\subsection{Perception: Instruments as Extensions of Scientific Senses}

Perception characterizes an agent's capacity to sense the scientific environment through instruments that extend the limits of human perception. In embodied science, instruments do not merely record data; they function as artificial scientific senses, defining what aspects of the physical world are observable and how experimental information is structured.
Scientific perception comprises two complementary forms of instrument sensing.

First, instruments provide instrument-mediated physical observations, through which latent physical phenomena are rendered observable. These include high-dimensional signals such as microscopy images, cryo-electron microscopy reconstructions, spectral measurements, and other modality-specific outputs. Such observations expose structural, dynamical, or compositional properties of experimental systems that are inaccessible to unaided human senses.
Second, instruments define instrument-defined experimental states, which encode the operational and procedural context of experimentation. These include progress indicators, equipment status, and structured records maintained in electronic lab notebooks or laboratory information systems. Rather than capturing physical phenomena directly, these signals formalize the evolving state of an experiment, enabling agents to track execution, detect deviations, and coordinate multi-step workflows.
Together, these two forms allow embodied agents to perceive not only what is occurring in an experiment, but also where it resides within the broader scientific process, forming a stable foundation for closed-loop reasoning and action.

\subsection{Language: Reasoning with Models, Knowledge, and Tools}
Language constitutes the ``scientific brain'' of an agent, responsible for scientific reasoning, interpretation, and planning. Within PLAD, this component is centered on large language models (LLMs)~\cite{naveed2025comprehensive,achiam2023gpt}, understood as a general class of foundation models that encompass multimodal LLMs~\cite{wang2024qwen2,zhuang2025advancing,bai2025interns1scientificmultimodalfoundation} capable of reasoning over heterogeneous scientific inputs.
While LLMs supply general reasoning capability, reliable scientific intelligence cannot emerge from models alone. Instead, it arises from a structured integration of models, specialized knowledge, and task-specific tools, enabling a dynamic balance between generality and specialization.

At the model level, LLMs used for scientific tasks must extend beyond natural language understanding to interpret multimodal scientific inputs produced by perception, including experimental data, instrument outputs, and structured records. In addition, the slow-thinking~\cite{guo2025deepseek} of LLMs enables sustained reasoning over complex instrument data as well as hypothesis inference and experimental planning. Such capacity allows models to integrate heterogeneous evidence, examine intermediate conclusions, and reason counterfactually about alternative explanations, experimental conditions, or mechanistic hypotheses.
These requirements motivate a set of science-oriented design considerations. These include modality-aware encoding for spectra, images, and time series; specialized tokenization schemes for chemical, biological, or materials representations; and architectural support for reasoning over long contexts. Training pipelines further benefit from integrating scientific literature with instrument-generated data, enabling LLMs to acquire long chain-of-thought reasoning patterns that align with empirical scientific practice.

Knowledge provides the specialized grounding and constraint that anchors general LLM reasoning in domain reality. Such knowledge spans unstructured scientific literature and structured resources, including databases and scientific knowledge graphs (Sci-KGs)~\cite{ding2025bridging}.
Sci-KGs systematically encode scientific concepts and their relationships in the form of triples, integrating structured databases with unstructured textual knowledge to provide a more comprehensive and stable foundation. Moreover, knowledge graphs can incorporate multimodal information, such as omics data, imaging data, and dynamical trajectories from computational simulations, thereby offering rich contextual support for scientific reasoning.
Importantly, grounding and constraint are not abstract properties but are operationalized through concrete mechanisms~\cite{meng2025integrating}. 
{During training, structured knowledge can be transformed into reasoning supervision, for example, via knowledge-graph-to-corpus approaches~\cite{DBLP:conf/naacl/AgarwalGSA21} that convert graph structure into long-chain scientific reasoning data, yielding high-reliability learning signals.}
During inference, literature, databases, and knowledge graphs are integrated through retrieval-augmented generation (RAG)~\cite{DBLP:journals/corr/abs-2312-10997,10.1145/3701716.3715240}, supplying authoritative context that constrains model outputs and stabilizes reasoning under uncertainty. In this way, knowledge injects precision, consistency, and domain depth that general-purpose LLMs alone cannot guarantee.

Tools constitute a second axis of specialization by extending reasoning into executable operations. 
{Through tool invocation, such as web search, database querying, and computational model invocation, LLMs can actively acquire external evidence, perform specialized analyses, and validate intermediate hypotheses.}
Unlike the general reasoning capacity of LLMs, tools encode expert procedures and formal methods that deliver accuracy and reliability on well-defined scientific tasks, effectively externalizing domain expertise into verifiable operations.

Together, this model–knowledge–tool integration enables a dynamic balance between generality and specialization. General-purpose LLMs provide adaptability, contextual understanding, and analogical reasoning across domains; specialized knowledge and tools deliver precision, depth, and task reliability. Within PLAD, agents dynamically adjust their reliance on these components according to task demands, allowing them to robustly interpret scientific data, perform mechanism-aware reasoning, and plan experiments that are both flexible across domains and dependable within specific scientific contexts.

\subsection{Action: Embodied Execution in Physical World}
Action corresponds to the scientific body of an embodied agent and denotes its capacity to intervene in the physical world through experimental execution. In PLAD, action is defined by the agent's ability to physically manipulate materials, instruments, and experimental processes, thereby grounding scientific reasoning in reality.
Embodied execution can be broadly organized along a key dimension: the degree of physical constraint under which an agent operates. Along this axis, embodied forms range from spatially constrained embodiments, which trade autonomy for reliability, to spatially unconstrained embodiments, which prioritize flexibility at the cost of execution certainty.

Spatially constrained embodiments operate within predefined mechanical boundaries and are tightly coupled to laboratory infrastructure, with two representative forms: stationary manipulators and linear track manipulators~\cite{lu2024automated}. Stationary manipulators are fixed robotic arms deployed at individual experimental stations, automating well-defined manual operations such as dispensing, sample loading or unloading, and instrument handling. Their limited work envelope enables high precision, stability, and repeatability, but also confines them to step-level execution within isolated experimental stages. Linear track manipulators extend this capability by mounting robotic arms on meter-scale rails that connect multiple functional stations, such as synthesis and characterization zones. This configuration enables coordinated, multi-step experimental workflows and sustained high-throughput execution under predefined pipelines, significantly expanding experimental coverage while preserving execution reliability. Nevertheless, both forms remain constrained by fixed layouts and preconfigured trajectories, favoring robustness and safety over autonomy and limiting their adaptability to unstructured or rapidly evolving laboratory settings.

In contrast, spatially unconstrained embodiments operate without predefined trajectories or fixed work envelopes. This category includes mobile manipulators, which integrate robotic arms with wheels, as well as humanoid robots~\cite{burger2020mobile,dai2024autonomous}. Such embodiments can navigate complex laboratory spaces, transport samples and consumables, and interact with diverse instruments across distributed environments. Their physical freedom enables higher-level autonomy and more closely mimics the behaviors of human researchers. At the same time, this flexibility introduces challenges in motion planning and execution reliability, particularly in safety-critical laboratory settings characterized by dense equipment and intricate procedures.

These two classes of embodiment define complementary regimes of embodied action~\cite{orouji2025autonomous,zhao2024biofoundry}. Spatially constrained systems provide a stable and reliable foundation for routine experimentation, while spatially unconstrained systems enable flexibility and integration across heterogeneous laboratory contexts. Among the latter, humanoid robots offer a distinctive long-term advantage: by directly operating within human-oriented laboratory layouts, tools, and protocols, they minimize the need for environment reconfiguration and thus represent a promising pathway toward general-purpose laboratory autonomy. Long-horizon autonomous scientific discovery arises from integrating constrained reliability with unconstrained adaptability, with embodied action in PLAD bridging reasoning and sustained physical experimentation.

\subsection{Discovery: Internalizing Execution Outcomes as Scientific Insight}
Discovery denotes the process by which an agent internalizes the outcomes of embodied action into scientific insight. Rather than treating experimental results as isolated observations, discovery abstracts execution feedback into transferable scientific understanding. Within PLAD, this process converts exploratory questions into new insights, such as refined interpretations of enzyme function, inferred reaction pathways, or updated structure–property relationships.  By feeding these insights back into subsequent reasoning and action, discovery enables the continual refinement of research objectives and supports long-horizon autonomous scientific discovery.

\begin{figure*}[t!]
    \centering
    \includegraphics[width=0.8\textwidth]{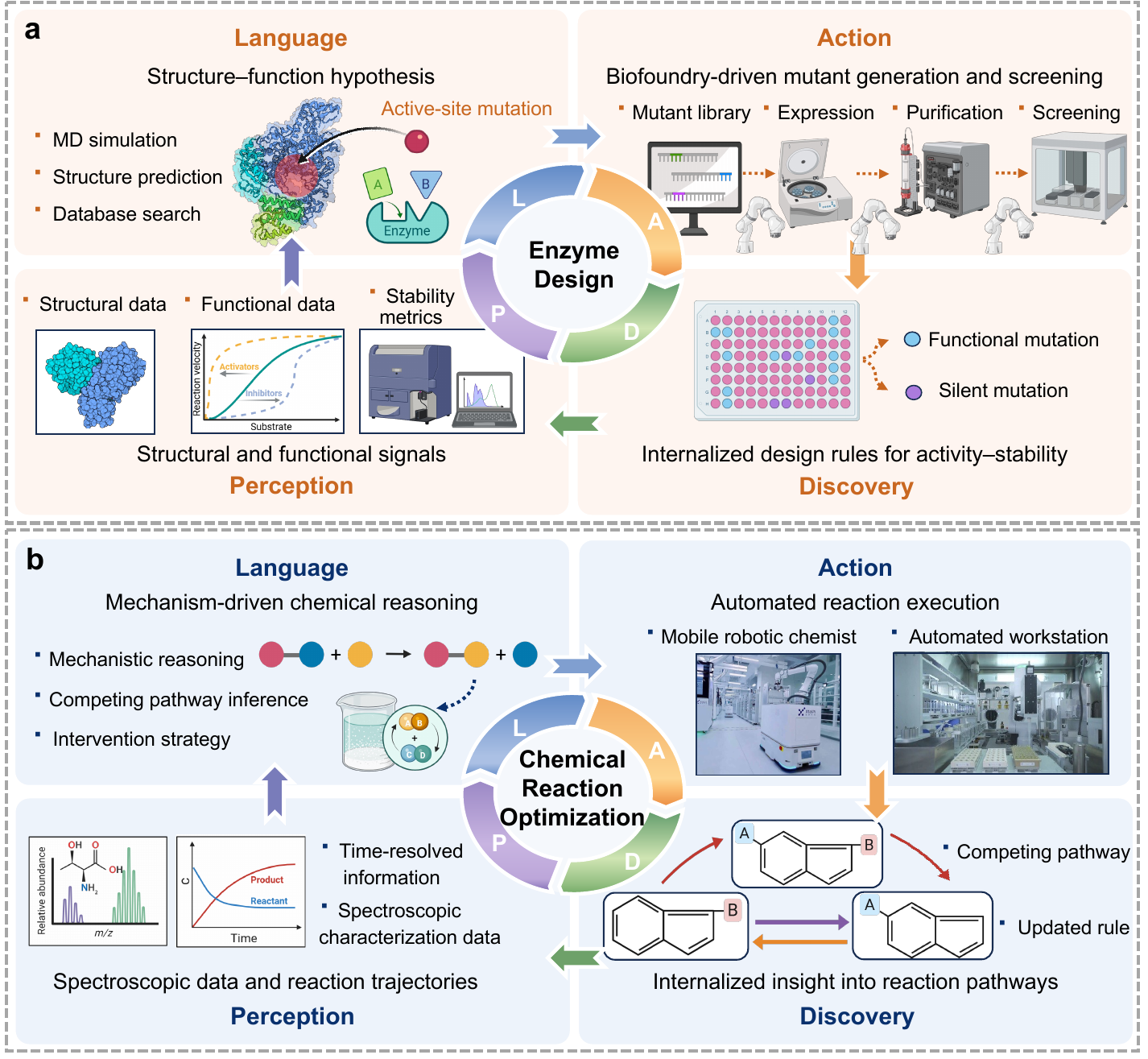}
    \caption{\textbf{Instantiating PLAD in scientific discovery.}
    \textbf{(a) Enzyme design:} Perception integrates structural and functional measurements; Language forms structure--function hypotheses under constraints; Action executes library construction and screening; Discovery internalizes design rules and counterexamples for future cycles.
    \textbf{(b) Chemical reaction optimization:} Perception tracks spectroscopic and kinetic trajectories; Language performs mechanism-aware reasoning; Action executes targeted campaigns; Discovery internalizes mechanistic explanations and pathway-level constraints that generalize beyond parameter tuning.}
    
    \label{fig:demo}
\end{figure*}

\subsection{PLAD Examples}
Using enzyme design and chemical reaction optimization as representative examples, this section illustrates how the PLAD framework can be instantiated across diverse experimental settings (Figure~\ref{fig:demo}). While the underlying scientific questions and experimental modalities differ, PLAD provides a common organizational structure for integrating perception, reasoning, and action into closed-loop discovery processes.

\subsubsection{Enzyme Design}
In enzyme design, Perception focuses on processing instrument-derived protein structural outputs (e.g., from cryo-electron microscopy or X-ray crystallography), functional and kinetic measurements (such as $K_m$, $k_{cat}$, and time-resolved activity profiles), and experimental indicators of stability and expression (including thermal stability, expression levels, and purification yield), among other inputs.
Language is centered on constructing hypotheses about enzyme structure–function relationships. It aims to infer which structural changes are likely to drive functional improvements. For example, it must determine whether enhanced activity arises from optimized substrate-binding conformations or from increased global stability, and distinguish whether activity loss results from direct perturbation of the active site or from conformational instability caused by distal mutations. Based on these inferences, it proposes design strategies, such as prioritizing mutations at active-site residues or shifting focus to modifying secondary-structure interfaces to enhance stability. This reasoning process is supported by computational and analytical tools, including protein structure prediction and molecular dynamics simulations to assess how mutations affect conformational stability and dynamics, as well as database queries to examine evolutionary conservation or homologous variant distributions.
Action subsequently translates these reasoning outcomes into concrete, embodied experimental execution. Action is implemented through control of physical experiments, such as generating mutant libraries on a biofoundry and orchestrating high-throughput expression, purification, and activity screening.
Through iterative cycles, structure–function hypotheses are tested and refined over extended experimental horizons.

\subsubsection{Chemical Reaction Optimization}
In chemical reaction optimization, Perception emphasizes dynamic and process-level signals, including spectroscopic characterization data (such as NMR and IR) as well as time-resolved information, including reaction progress curves and by-product formation trajectories. Together, these signals reflect the temporal evolution of reaction pathways, intermediate states, and operational stability.
Language is centered on mechanism-driven chemical reasoning. It focuses on how solvent, additives, and ligand structure modulate key elementary steps, such as oxidative addition, migratory insertion, and reductive elimination. When undesired outcomes arise, such as diminished enantioselectivity or increased by-product formation, the focus shifts from global parameter adjustment to targeted hypothesis refinement. The formation of specific side products may indicate a competing mechanistic pathway, prompting structural modifications or the introduction of additives to suppress or intercept that pathway. This reasoning process is also supported by computational tools, including quantum chemical modeling and kinetic analysis, which are used to assess relative energetics, barrier heights, and selectivity trends under different conditions.
Action then translates these into embodied experimental execution. Actions are realized through automated workstations or mobile robotic chemists that execute targeted reaction campaigns, experimentally validating mechanistic hypotheses and suppressing undesired pathways. Through such embodied intervention, mechanism-aware exploration can be sustained across extended experimental cycles.
\begin{figure*}[t!]
    \centering
    \includegraphics[width=\textwidth]{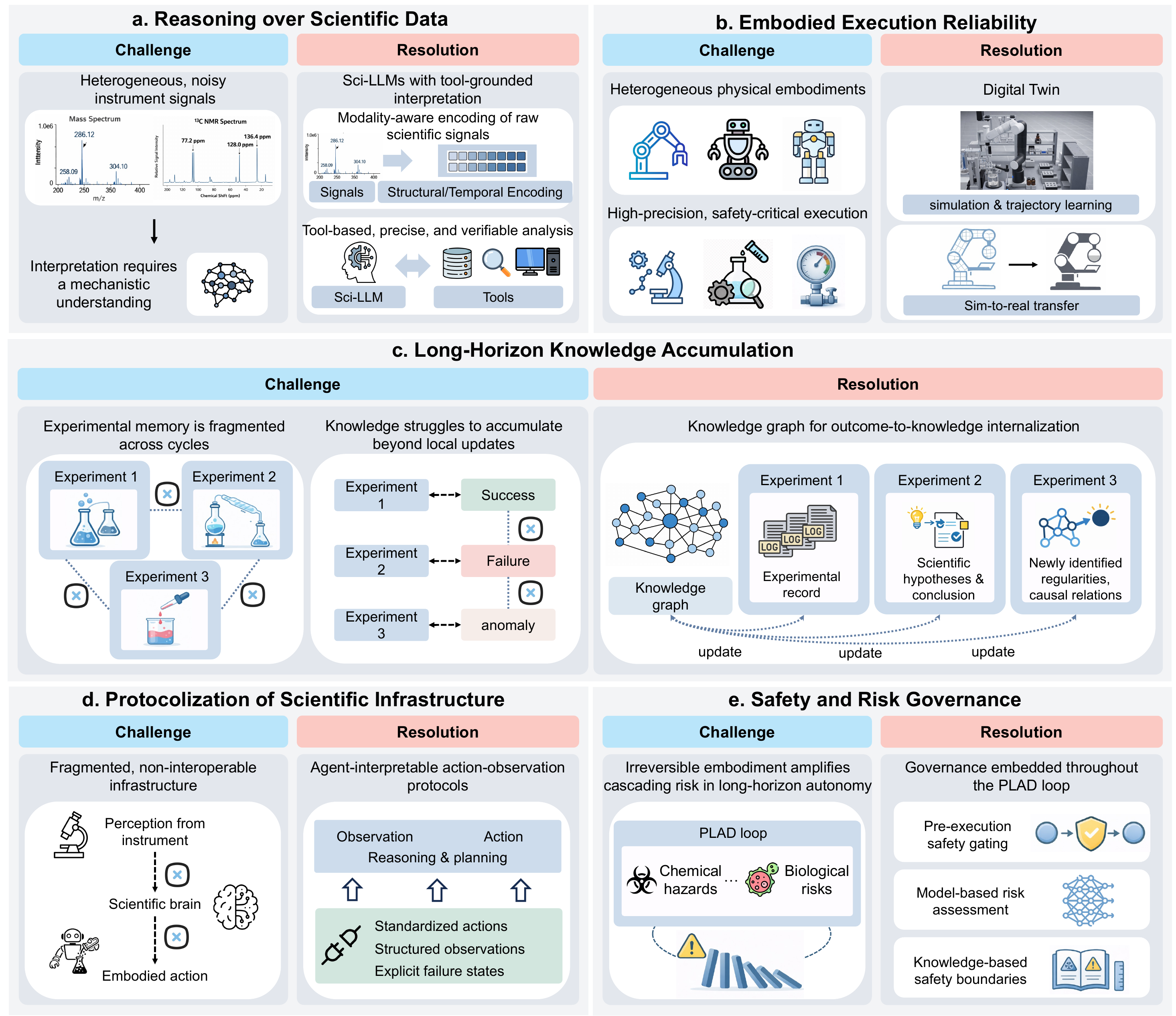}
    \caption{
    \textbf{Challenges and design resolution for sustaining long-horizon PLAD loops}.
(a) \textbf{Reasoning over scientific data}: Heterogeneous, noisy instrument signals require mechanism-aware interpretation; science-adapted LLMs, modality-aware encoders, and tool-grounded analysis support reliable reasoning.
(b) \textbf{Embodied execution reliability}: Robust execution across diverse robotic embodiments requires large-scale simulation, trajectory learning, and sim-to-real transfer.
(c) \textbf{Long-horizon knowledge accumulation}: Knowledge graphs convert fragmented experimental records into cumulative scientific insight across cycles.
(d) \textbf{Protocolization of scientific infrastructure}: Agent-interpretable action–observation protocols standardize actions, observations, and failure states, enabling composable closed-loop workflows.
(e) \textbf{Safety and risk governance}: Trustworthy long-horizon autonomy requires integrated safety gating, model-based risk assessment, and knowledge-based safety boundaries.
    }
    
    \label{fig:cha}
\end{figure*}
\section{Challenges and Design Considerations}
In this section, we identify challenges that constrain the stability, reliability, and safety of long-horizon PLAD loops, and outline corresponding design responses that are necessary to keep such systems operational, cumulative, and trustworthy (Figure~\ref{fig:cha}).

\subsection{Reasoning over Scientific Data}
Reasoning over scientific data is fundamental to bridging the gap between Perception and Language. 
{In experiments, perceptual inputs typically manifest as complex, noisy signals, such as liquid chromatography–mass spectrometry (LC–MS) spectra, nuclear magnetic resonance (NMR) and infrared (IR) spectra, time-resolved reaction kinetic profiles, microscopy images, and instrument status logs.}
Interpretation of these data is deeply contingent on domain-specific knowledge and contextual experimental details.
Consider LC–MS data as an illustrative example: a spectrum is not a simple “product fingerprint” but rather the superposition of multiple physicochemical processes, including variations in ionization efficiency, matrix effects, diverse fragmentation pathways, co-elution of isomers, and dynamic changes in signal-to-noise ratio over time. Consequently, the appearance, disappearance, or intensity shift of a peak does not necessarily correlate with reaction progress or the formation of a target compound; accurate interpretation requires integrating retention time, isotopic distribution patterns, fragment ion signatures, and experimental conditions into a coherent assessment.
Similarly, scientific instruments characterize reaction through process signals and operational states. 
{Parameters such as temperature, pressure, and flow rate, along with sensor readings or visual cues (e.g., phase transitions or color evolution captured in images), are routinely used to assess whether a reaction has stalled or deviated from its intended trajectory.}
Such judgments, however, also rely on a deep understanding of the underlying reaction mechanisms and experimental protocols.

Addressing this challenge requires coordinated advances in model design and tool-assisted reasoning. At the architectural level, large language models must be adapted to scientific environments so as to accommodate the structural or temporal properties of experimental data. This enables reasoning to operate directly over instrument-generated representations, rather than relying solely on linguistic descriptions. Such science-adapted language models are often described as scientific large language models (Sci-LLMs)~\cite{bai2025interns1scientificmultimodalfoundation}.
Even with these adaptations, it is difficult for LLMs to natively cover all dimensions of scientific data analysis. Reliable interpretation also depends on tool invocation, in which models work in concert with specialized computational tools to execute precise analytical procedures while maintaining control over higher-level scientific inference.
Both model-based and tool-assisted reasoning rely critically on appropriate training paradigms. Agentic reinforcement learning has been shown to be effective in training slow-thinking and contextual tool use, enabling agents to invoke, interpret, and integrate tool outputs within sustained reasoning and planning processes. Together, these design and training strategies support the analysis of raw experimental signals within long-horizon autonomous discovery loops.

\subsection{Embodied Execution Reliability}
The core challenge of embodied execution in scientific scenarios lies in execution reliability: whether the experimental actions can be executed correctly, consistently, and repeatedly in complex laboratory environments under strict safety constraints.
Scientific experimentation may not be carried out by a single, uniform robotic form. Instead, embodied execution spans a diverse set of physical carriers, including stationary or mobile manipulators and humanoid robots. Ensuring reliable execution across such heterogeneous embodiments amplifies the difficulty of autonomy in scientific environments.
{Beyond the diversity of embodiments, scientific experiments themselves demand a wide range of precise and highly specialized skills, such as liquid dispensing, powder weighing, apparatus grasping, sample transfer, and instrument interfacing.}
More importantly, real experiments often involve hazardous chemicals, high temperatures, high pressures, or biosafety risks, making it difficult to scale up learning approaches that rely on physical trial and error.

To address these challenges, sim-to-real approaches based on digital twin environments play a central role~\cite{zhao2020sim}. Digital twins enable systematic modeling not only of laboratory layouts and instrument geometries, but also of the interaction dynamics specific to different embodiments and experimental mechanisms.
In scientific scenarios, the key to digital twins is not limited to simulation accuracy in terms of geometry or motion, but also includes the accurate depiction of experimental mechanisms. For example, by modeling heat transfer processes, temperature changes in the virtual environment can accurately reflect thermal behaviors in physical experiments, ensuring that execution strategies developed in the simulation environment remain valid in real experiments~\cite{darvish2025matterix}. Leveraging simulation environments for large-scale, low-cost training and trial and error, combined with high-quality trajectory data generated from human demonstrations and robot executions, can gradually narrow the gap between virtual and real worlds, enabling smooth migration from simulation training to real-world deployment.

\subsection{Long-Horizon Knowledge Accumulation}
The essence of long-horizon autonomy lies in the sustained bidirectional closure between cognition and execution across multiple experimental cycles. Scientific discovery is not a collection of isolated experiments, but a process of cumulative knowledge construction, revision, and extension.
First, long-horizon autonomy imposes stringent requirements on memory management~\cite{hu2025memory}. Agents must continuously record, organize, and revisit historical hypotheses, experimental designs, and observations across extended timescales, such that prior experience can be reliably carried forward into future decision-making. Such continuity cannot be reliably supported by language-model context windows or unstructured experimental logs alone.
Second, long-horizon discovery requires that experimental outcomes be systematically analyzed and internalized as evolving scientific knowledge. Newly generated results should not remain as transient observations or isolated performance metrics; rather, they must be integrated into the agent's internal state. This integration ensures that past successes, failures, and anomalies exert lasting influence, thereby enabling cumulative rather than repetitive discovery.

A central strategy for addressing these challenges is the introduction of knowledge graphs as structured representational skeletons~\cite{chhikara2025mem0}. At the memory level, they transform fragmented experimental records, hypotheses, and conclusions into structured representations that support retrieval, comparison, and reasoning across temporal scales. At the discovery level, newly identified scientific regularities, causal relationships, or anomalous behaviors can be incorporated as new nodes or relationships, allowing scientific understanding to be continuously expanded and refined.
Overall, long-horizon knowledge accumulation is a systemic requirement for the stability and effectiveness of the entire closed loop. Only when scientific knowledge can be persistently accumulated, structured, and revised across experimental cycles can embodied agents move beyond short-term optimization and genuinely assume responsibility for long-horizon autonomous scientific discovery.

\subsection{Protocolization of Scientific Infrastructure}
In most laboratories, experimental devices, sensing instruments, and execution modules typically operate as isolated systems. Scientific instruments extend the perceptual boundary of discovery by rendering processes observable; their states and outputs need to be acquired continuously to support reasoning. In practice, however, measurements, device states, and operational logs are recorded locally or exposed only as low-level signals. As a result, perceptual information cannot flow continuously into the scientific brain, and reasoning outcomes cannot be reliably grounded in the evolving experimental state.
This challenge is further amplified in complex experimental settings that involve multi-step procedures and coordinated embodied action. Executing such workflows requires agents to orchestrate heterogeneous robots and manipulators. Without a unified representation of actions and system states, high-level experimental plans cannot be systematically decomposed into executable embodied behaviors. Together, these factors create a structural disconnection between perception, reasoning, and action, constituting a fundamental bottleneck for closed-loop autonomy.

Overcoming this bottleneck requires a principled reorganization of how experimental components are represented and interfaced.
Protocolized infrastructure addresses this challenge by defining a unified, agent-interpretable abstraction over distributed experimental components~\cite{jiang2025scp,sim2024chemos}. While networked connectivity enables heterogeneous instruments, sensors, and execution systems to expose their states and outputs, protocolization lifts raw connectivity into agent-operable capability.
For example, the Science Context Protocol (SCP)~\cite{jiang2025scp} offers a standardized way to expose and orchestrate scientific resources, including tools, models, datasets, and physical instruments, thereby transforming heterogeneous laboratory connectivity into agent-operable capability.
By standardizing how experimental actions, perceptual observations, and failure or exception states are represented, protocols enable agents to interpret, compare, and compose interactions across instruments and experimental contexts.
This allows actions and observations to function as reusable, verifiable primitives within the PLAD loop, supporting traceability, composability, and cumulative discovery across long-horizon cycles.

\subsection{Safety and Risk Governance}
Safety constitutes a fundamental challenge for long-horizon autonomous scientific exploration. A defining feature of PLAD is that experimental actions are \emph{embodied} and \emph{irreversible}. Physical interventions permanently alter materials, instruments, and environmental states, and erroneous actions cannot be rolled back as failed computations. Risk is further amplified by the iteration of PLAD. It also creates the potential for \textit{runaway autonomy}, progressively relax implicit safety margins. These factors substantially amplify safety hazards during experimentation, including the execution of unsafe procedures or the generation of hazardous outcomes. For example, operating under extreme temperatures or pressures, employing toxic reagents outside validated regimes, or producing products that pose chemical, biological, or environmental risks beyond the experimental boundary.

Safety governance in PLAD relies on a deliberate complementarity between knowledge-driven constraints and model-based risk assessment. Explicit safety knowledge can be derived from laboratory regulations, instrument specifications, hazard databases, and experimental manuals. It defines operational boundaries that delimit where autonomous exploration must not go. These constraints encode known hazards and non-negotiable limits, ensuring that hypothesis generation and experimental planning remain within validated and auditable regimes. However, long-horizon autonomous discovery routinely encounters risk that cannot be exhaustively specified in advance. As PLAD iterates, hazards may emerge from contextual interactions, cumulative deviations, or gradual extrapolation toward extreme conditions. Model-based safety supervision is therefore required to assess such context-dependent and emergent risks. Safety-aware guard models evaluate candidate plans within their historical and experimental context, estimating whether sequences of otherwise permissible actions collectively approach unsafe regimes.
\section{Conclusion}
Embodied Science reframes discovery as a long-horizon closed-loop process in which reasoning is continuously grounded, corrected, and enriched through physical interaction with the world.
{This perspective highlights a structural limitation in today’s AI4S landscape: scaling language reasoning improves cognitive breadth, and advancing laboratory automation improves throughput, but neither alone satisfies the core requirement of autonomous discovery, namely the ability to iteratively generate, test, falsify, and revise hypotheses over extended horizons.}
Without embodiment, reasoning risks becoming self-referential; without scientific cognition, execution risks degenerating into blind optimization.

PLAD provides an organizing principle for overcoming this divide.
{By integrating Perception, Language, Action, and Discovery into a coupled system, PLAD shifts autonomy from short-term performance to cumulative scientific understanding, learning not only from success but also from failure, anomaly, and uncertainty.}
Realizing this vision will require coordinated advances across foundation models, instrument-aware perception, protocol compilation and control, scientific infrastructure, evaluation standards, and safety governance.
The central question is no longer whether AI can assist science, but whether scientific discovery can be engineered as an embodied, long-horizon process that remains trustworthy as autonomy scales.
\bibliography{ref}
\end{document}